\pdfoutput=1
\documentclass[11pt]{article}

\usepackage[final]{EACL2023}

\usepackage{times}
\usepackage{latexsym}
\usepackage[whole]{bxcjkjatype} 
\usepackage{graphicx,xcolor} 
\usepackage{tabularx}
\usepackage{amsmath,amssymb}
\usepackage{mathtools}
\usepackage{here}
\usepackage{multirow}
\usepackage{booktabs,graphicx,makecell,siunitx,eqparbox}
\usepackage{booktabs}
\usepackage{subcaption}
\usepackage[T1]{fontenc}
\usepackage[utf8]{inputenc}
\usepackage{microtype}
\usepackage{inconsolata}

\title{Empirical Investigation of Neural Symbolic Reasoning Strategies}
\author{Yoichi Aoki,${}^{1}$ 
        Keito Kudo,${}^{1}$
        Tatsuki Kuribayashi,${}^{1,2}$
        Ana Brassard,${}^{3,1}$ \\ 
        {\bf Masashi Yoshikawa,${}^{1}$ Keisuke Sakaguchi,${}^{1,3}$ Kentaro Inui${}^{1,3}$} \\
         ${}^{1}$Tohoku University,
         ${}^{2}$Langsmith, Inc.,
         ${}^{3}$RIKEN \\ 
        \texttt{\{youichi.aoki.p2,keito.kudo.q4\}@dc.tohoku.ac.jp, } \\
        \texttt{kuribayashi@tohoku.ac.jp, } 
        \texttt{ana.brassard@riken.jp, } \\
        \texttt{\{yoshikawa,keisuke.sakaguchi,kentaro.inui\}@tohoku.ac.jp}}

\definecolor{outputblue}{RGB}{35, 128, 198}
\definecolor{chainingpink}{RGB}{221, 63, 128}
\begin{document}
\maketitle
%%%%%%%%%%%%%%%%%%%%%%%%%%%%%%%%%%%%%%
\newcommand{\neuralmodel}{neural NLP models}
%%%%%%%%%%%%%%%%%%%%%%%%%%%%%%%%%%%%%%
\newcommand{\multihop}{multi-hop inference}
\newcommand{\numerical}{numerical inference}
\newcommand{\multihopnumerical}{numerical multi-hop inference }
%%%%%%%%%%%%%%%%%%%%%%%%%%%%%%%%%%%%%%
\newcommand{\inferenceprocess}{inference processes}
\newcommand{\outputmethod}{output methods}
\newcommand{\searchstrategy}{inference paths}
%%%%%%%%%%%%%%%%%%%%%%%%%%%%%%%%%%%%%%
\newcommand{\forwardall}{exhaustive path}
\newcommand{\forwardlimit}{shortest forward path}
\newcommand{\backward}{backward path}
\newcommand{\decoderall}{decode-all}
\newcommand{\decoderstep}{decode-step}
\newcommand{\decodertoken}{decode-char}
%%%%%%%%%%%%%%%%%%%%%%%%%%%%%%%%%%%%%%
\begin{abstract}
\label{abstract}
Neural reasoning accuracy improves when generating intermediate reasoning steps. However, the source of this improvement is yet unclear. Here, we investigate and factorize the benefit of generating intermediate steps for symbolic reasoning.
Specifically, we decompose the reasoning strategy w.r.t. step granularity and chaining strategy. With a purely symbolic numerical reasoning dataset (e.g., \texttt{A=1, B=3, C=A+3, C?}), we found that the choice of reasoning strategies significantly affects the performance, with the gap becoming even larger as the extrapolation length becomes longer. Surprisingly, we also found that certain configurations lead to nearly perfect performance, even in the case of length extrapolation. Our results indicate the importance of exploring effective strategies for neural reasoning models.~\footnote{Code available at:  \url{https://github.com/ao1neko/reasoning-strategy}}
\end{abstract}
\section{Introduction}
\label{introduction}
Artificial intelligence researchers have been attempting neural-symbolic integration for a long time~\cite{Garcez2020,Hamilton2022}.
Neural models tend to perform better when generating intermediate reasoning steps in addition to the answer.
This phenomenon was seen across various reasoning tasks, such as math word problems~\cite{Wei2022,Cobbe2021,Kojima2022,Recchia2021,Lewkowycz2022}, commonsense reasoning~\cite{Wei2022,Wang2022}, and symbolic reasoning~\cite{Wei2022,Kojima2022}.
However, it is yet unclear which factors in the intermediate step generation bring the benefit. 
Previous studies often used different strategies for step generation in an ad-hoc manner. 
%%%%%%%%%%%%%%%%%%%%%%%%%%%%%%%%%%%%%%%%%%%%%%
\begin{figure}
\centering
\includegraphics[width=\linewidth]{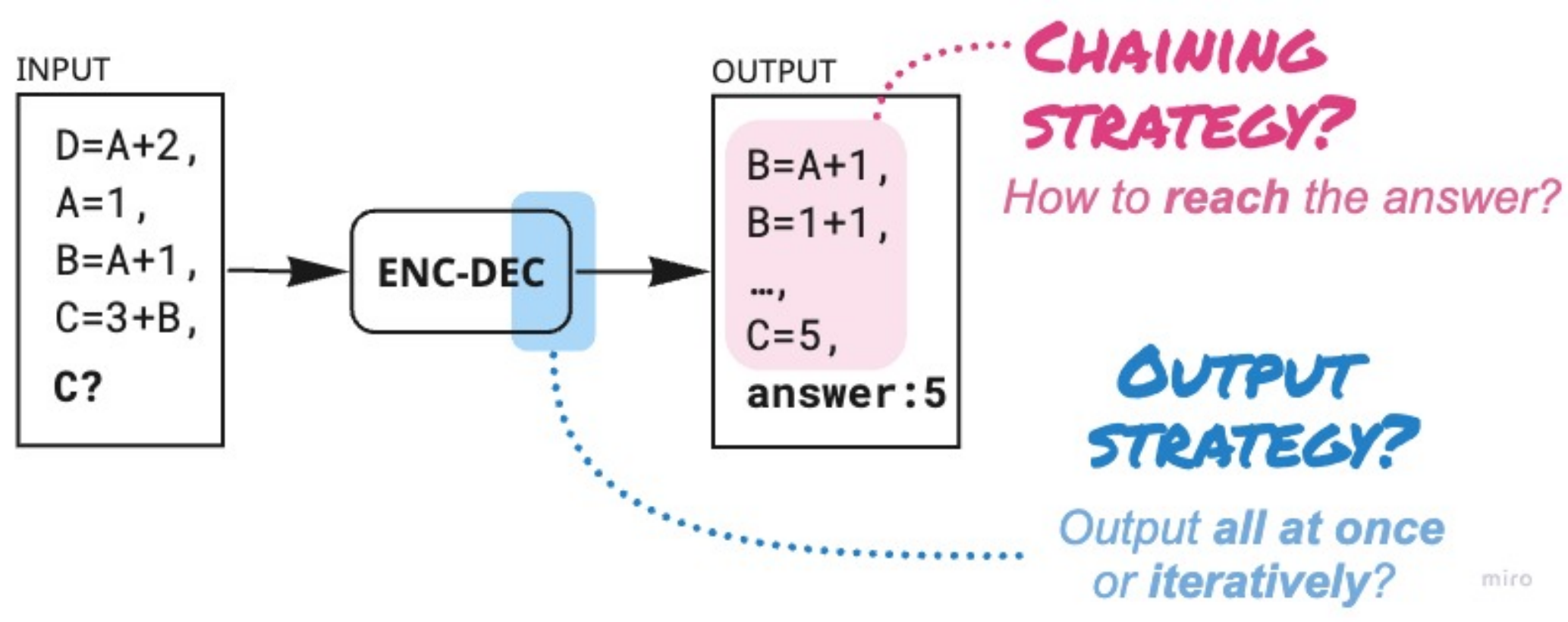}
\caption{In a controlled setting, we found that output and chaining strategy choice significantly impact performance when conducting multi-step reasoning.}
\label{fig:figure1}
\end{figure}
%%%%%%%%%%%%%%%%%%%%%%%%%%%%%%%%%%%%%%%%%%%%%%
%%%%%%%%%%%%%%%%%%%%%%%%%%%%%%%%%%%%%%%%%%%%%%
\begin{figure*}
    \centering
    \begin{subfigure}[c]{0.43\textwidth}
        \includegraphics[width=\textwidth]{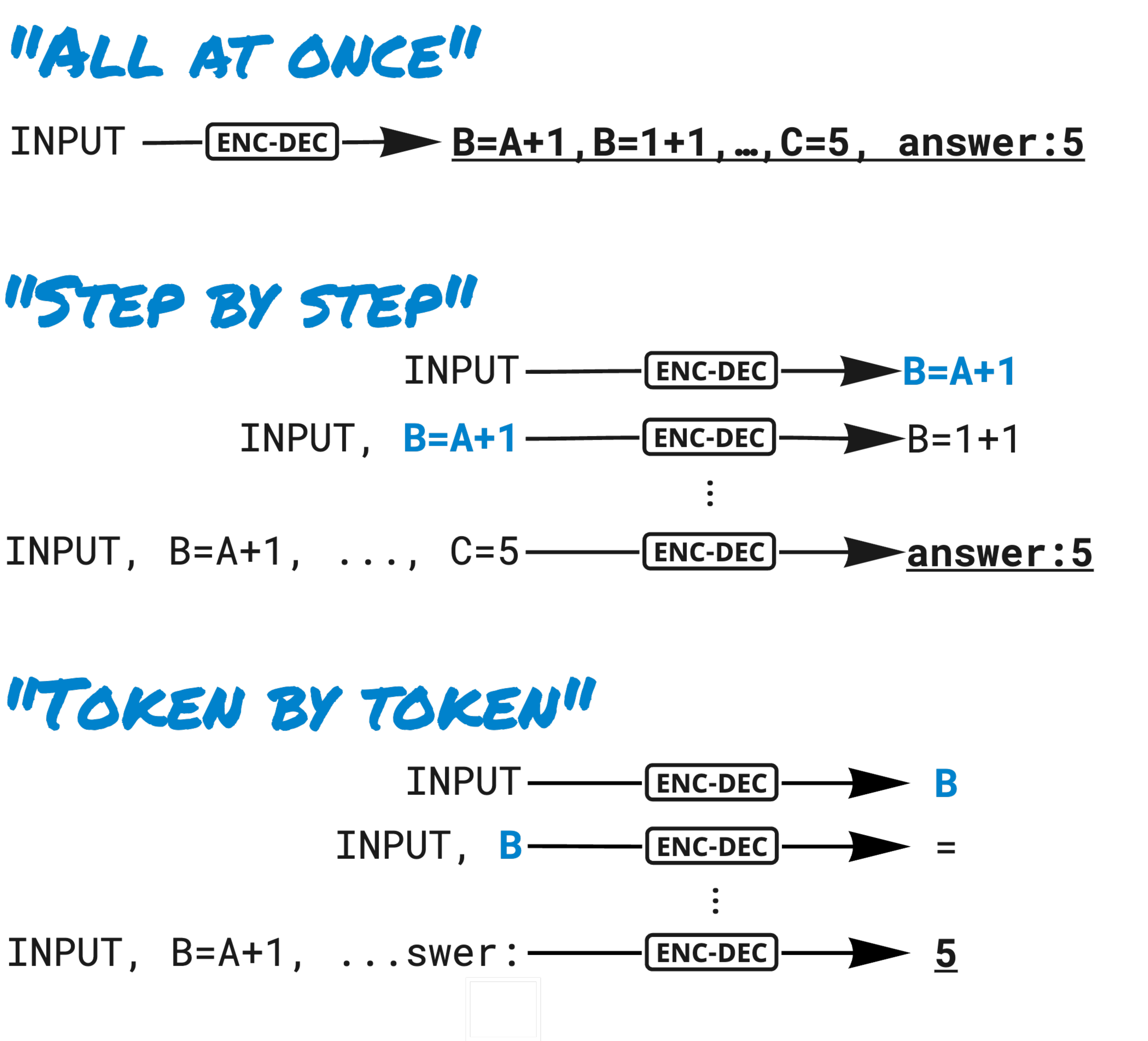}
        \caption{\emph{All-at-once}: output the entire reasoning chain and answer in a single call. 
        \emph{Step-by-step}: iteratively build the output with a single calculation step per call. 
        \emph{Token-by-token}: iteratively output only one \emph{token} per call.}
        \label{fig:output-strategies}
    \end{subfigure}
    \hfill
    \begin{subfigure}[c]{0.53\textwidth}
        \includegraphics[width=\textwidth]{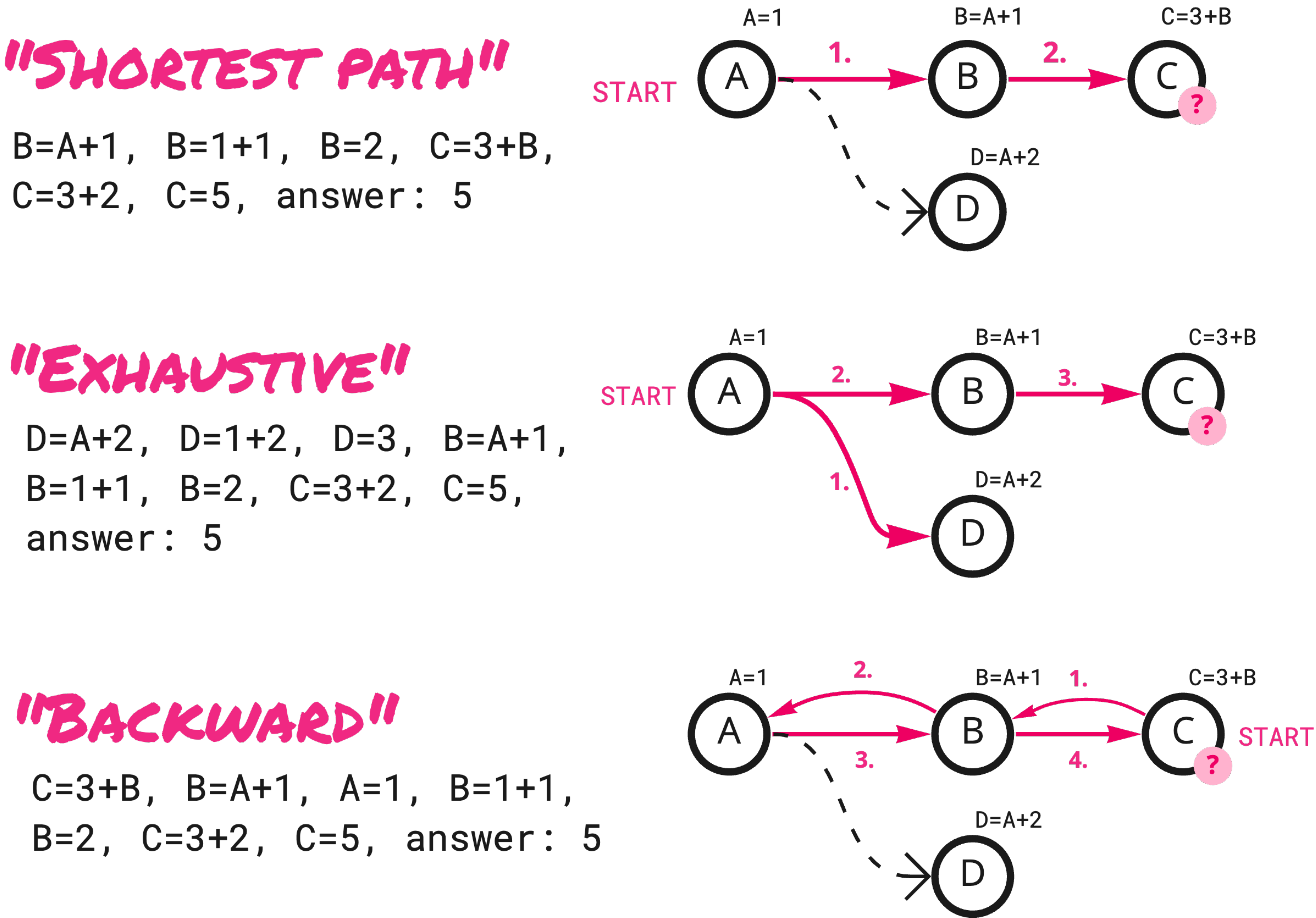}
        \caption{The graph nodes represent variables and edges their dependencies. \emph{Shortest path}: a minimal chain starting from the first necessary equation.
        \emph{Exhaustive}: greedily solve all equations until the target is reached. 
        \emph{Backward}: start from the target's equation, backtrack over dependencies until a known value is reached, then solve each equation in order.}
        \label{fig:chaining-strategies}
    \end{subfigure}
    \caption{Overview of (a) output and (b) chaining strategies given the  \texttt{INPUT}: \texttt{D=A+2, A=1, B=A+1, C=3+B, C?}}
    \label{fig:figure2-details}
\end{figure*}
%%%%%%%%%%%%%%%%%%%%%%%%%%%%%%%%%%%%%%%%%%%%%%
To investigate this, we break down the neural reasoning process into two strategies: \emph{output strategy} and \emph{chaining strategy} (Figure~\ref{fig:figure1}).
The output strategy (\S\ref{decoding_methods}) determines the granularity of intermediate reasoning step generation (all at once vs. step-by-step vs. token-by-token). Some studies trained the models to generate reasoning steps and a conclusion derived from them at once~\cite{Nye2021,Lewkowycz2022,Wei2022,Kojima2022,Wang2022,Recchia2021}, some generated 
a single reasoning step given the input and iterated this process until achieving a conclusion~\cite{Sanyal2022,Picco2021,Tafjord2021}, and others iteratively generated sub-goals as well as reasoning steps~\cite{Liang2021,Shwartz2020}.

In turn, the chaining strategy (\S\ref{reasoning_step}) defines the reasoning path direction (shortest path vs. exhaustive path vs. backward path). For example, some studies used a backward chaining process~\cite{Picco2021,DBLP:conf/nips/Rocktaschel017,DBLP:conf/aaaiss/CingilliogluR19}, while others adopted exhaustive searches~\cite{Tafjord2021,Liang2021,DBLP:journals/corr/abs-2205-12443}.

To compare the strategies, we prepared a test bed of numerical reasoning problems in a simplified language (Figure~\ref{fig:figure1}). This format allows for more controlled testing while serving as a necessary condition---should a model fail to solve it, it cannot be expected to adequately generalize to more complex math word problems.

We found that both strategies substantially affect the symbolic reasoning performance of neural seq2seq learners. 
Overall, iterative generation outperformed all-at-once outputting, and roughly granular reasoning steps (i.e., shortest-path chaining) lagged behind finely granular steps (i.e., exhaustive and backward chaining). 
Surprisingly, some settings had near-perfect performance even in generalization tests which extrapolate over greater reasoning depths and unseen numbers during training.
%%%%%%%%%%%%%%%%%%%%%%%%%%%%%%%%%%%%%%%%%%%%%%%%
\section{Experimental settings}
\paragraph{Problem definition.}
%\section{Problem definition}
\label{task}
We evaluated the models' ability to iteratively perform arithmetic operations over given symbols.
Given a series of equations, the task is to answer the value of a target variable (Figure~\ref{fig:figure1}).
Each question also has a certain reasoning depth---the number of \emph{necessary} equations to reach the answer. 
For example, the depth of the question \texttt{A=1, B=2+A, C=3+B, D=2, C?} is $3$ (\texttt{A=1, B=2+A, C=3+B}). 

Each equation defines either an assignment (e.g., \texttt{A=1}) or a modular addition and an assignment (e.g., \texttt{B=3+1}). The addition is mod 100.
The question contexts also contain distractors that are not necessary to calculate the answer (e.g., \texttt{D=A+2} in Figure~\ref{fig:figure1}).
A value assigned to a particular variable is typically referred to in different equations (e.g., \texttt{A=1, B=A+1}).
Numbers, variables, and the ordering of equations are randomly assigned.

\paragraph{Motivation for using artificial data}
There are mainly three advantages to this dataset. First, the symbolic format allows easier control of reasoning depth for generalization tests. %our setting is easier to control the reasoning depths for the generalization test by using the purely symbolic format.
Specifically, we trained a model using instances with shallow (1-5) depths and evaluated them with instances with shallow/deep (1-12) depths. On the other hand, math word problems are harder to control for reasoning depth (e.g., it is not easy to come up with various instances which have a reasoning depth of 10). 
Second, we wanted to avoid the "spurious bias" that natural (math word) texts implicitly bring into the model~\cite{Gururangan2018,Gupta2021,AI-Negheimish2021,Sugawara2018,Jia2017,McCoy2019}.
Third, we assume that our setting is the necessary condition for solving math word problems. It is unreasonable to expect that a model that can't solve this pure numerical reasoning task can solve more complex tasks.

\label{settings}
In total, we prepared 5K instances for training and 2.4K for testing.
%%%%%%%%%%%%%%%%%%%%%%%%%%%%%%%%%%%%%%%%%%%%%%%%
\subsection{Output strategies}
\label{decoding_methods}
\label{models}
We compared three configurations: all-at-once, step-by-step, and token-by-token (Figure~\ref{fig:output-strategies}).
%%%%%%%%%%%%%%%%%%%%%%%%%%%%%%%%%%%%%%%%%%%%%%%%

\noindent
\textbf{All-at-once: }
The model outputs the entire reasoning chain and the final answer in a single call (i.e., \textit{chain-of-thought} style)~\cite{Wei2022,Cobbe2021,Yavuz2022,Shwartz2020} .
In this setting, the more reasoning steps, the longer the sequence the decoder must generate at once.
%%%%%%%%%%%%%%%%%%%%%%%%%%%%%%%%%%%%%%%%%%%%%%%%

\noindent
\textbf{Step-by-step: }
The model outputs a single reasoning step per call. 
Each generated step is concatenated to the past input, and the model again generates the next step (i.e., \textit{proofwriter} style)~\cite{Liang2021,Sanyal2022,Picco2021,Tafjord2021,Shwartz2020} .
This process is iterated until the model outputs the answer or until a set maximum number of iterations is reached ($100$). 
%%%%%%%%%%%%%%%%%%%%%%%%%%%%%%%%%%%%%%%%%%%%%%%%
\noindent
\textbf{Token-by-token: }
This is the same as step-be-step chaining, but the decoder outputs only a single \emph{token} per call.
We set the maximum number of steps to $500$. 

Comparing \textit{all-at-once}\ and the others reveals the effect of changing the sequence length that the decoder outputs in a single call.
In addition, comparing \textit{step-by-step}\ and \textit{token-by-token}\ quantifies the advantage of breaking a problem into meaningful units.
%%%%%%%%%%%%%%%%%%%%%%%%%%%%%%%%%%%%%%%%%%%%%%%%
\subsection{Chaining strategies}
\label{reasoning_step}
\label{inference processes}
Particular variables sometimes depend on another variable; the key to reaching the correct answer is determining the order in which the equations are referred to.
Regarding existing studies, we compared three chaining strategies:
\emph{shortest-path}, \emph{exhaustive}, and \emph{backward} chaining
(Figure~\ref{fig:chaining-strategies}). 
%%%%%%%%%%%%%%%%%%%%%%%%%%%%%%%%%%%%%%

\noindent
\textbf{Shortest-path chaining: }
The model straight-forwardly solves the equations starting from the first solvable one (i.e., involving a known value) and ending with the target~\cite{Wei2022,Cobbe2021,Yavuz2022,Shwartz2020}.
Here, the reasoning behind determining the shortest path is not outputted by the model.
%%%%%%%%%%%%%%%%%%%%%%%%%%%%%%%%%%%%%%%%%%%%%%%%

\noindent
\textbf{Exhaustive chaining: }
The model greedily solves all given equations until the target value is reached~\cite{Tafjord2021,Liang2021,DBLP:journals/corr/abs-2205-12443}.
Specifically, the model calculates the left-most solvable equation in each step.
Note that this strategy typically derives a long reasoning chain; from an engineering perspective, this strategy is inefficient.
%%%%%%%%%%%%%%%%%%%%%%%%%%%%%%%%%%%%%%%%%%%%%%%%

\noindent
\textbf{Backward chaining: }
The model starts from the equation for the target variable and backtracks over the dependent equations until it reaches a known value~\cite{Picco2021,DBLP:conf/nips/Rocktaschel017,DBLP:conf/aaaiss/CingilliogluR19}.
Then, it solves each equation in order by inserting known or calculated values until the target one is reached. 
%%%%%%%%%%%%%%%%%%%%%%%%%%%%%%%%%%%%%%%%%%%%%%%%

\noindent
\textbf{No chaining: }
As a baseline, we also examined the setting where the model was trained to directly output the answer.
%%%%%%%%%%%%%%%%%%%%%%%%%%%%%%%%%%%%%%%%%%%%%%%%
\begin{figure}[t]
\centering
\includegraphics[width=7.5cm]{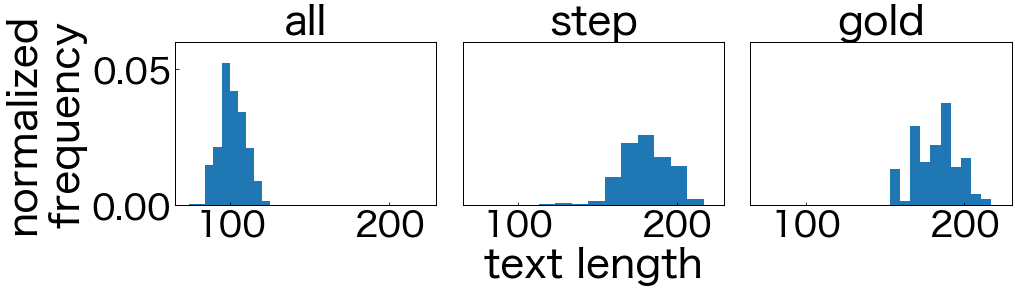}
\caption{Distributions of the total reasoning chain length (num. characters). The all-at-once and step-by-step generate those at depth 12.}
\label{fig:hist}
\end{figure}
%%%%%%%%%%%%%%%%%%%%%%%%%%%%%%%%%%%%%%%%%%%%%%%%
\begin{figure*}
    \centering
    \begin{subfigure}[c]{0.48\textwidth}
        \includegraphics[width=\textwidth]{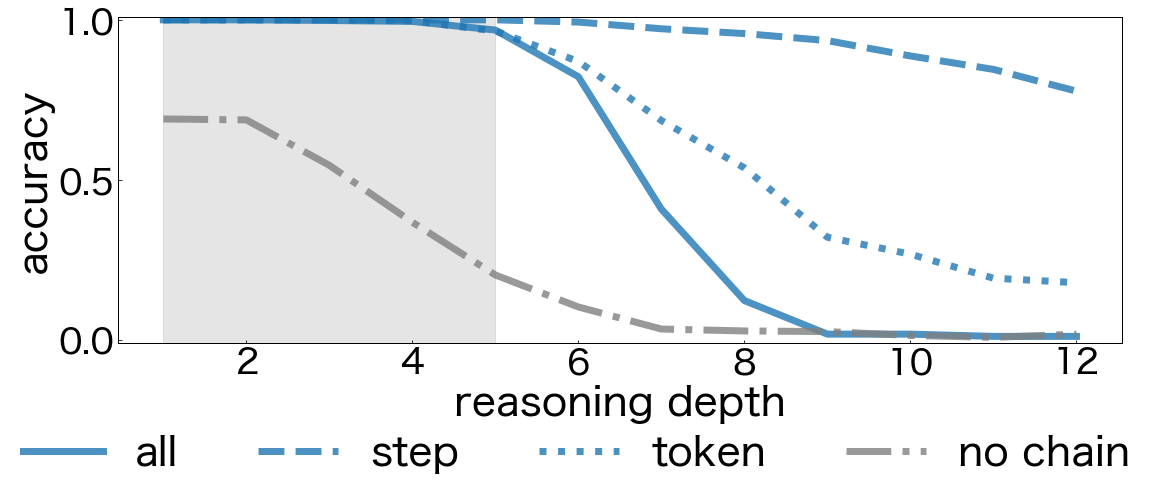}
\caption{Output strategy}
        \label{fig:result-output}
    \end{subfigure}
    \hfill
    \begin{subfigure}[c]{0.48\textwidth}
        \includegraphics[width=\textwidth]{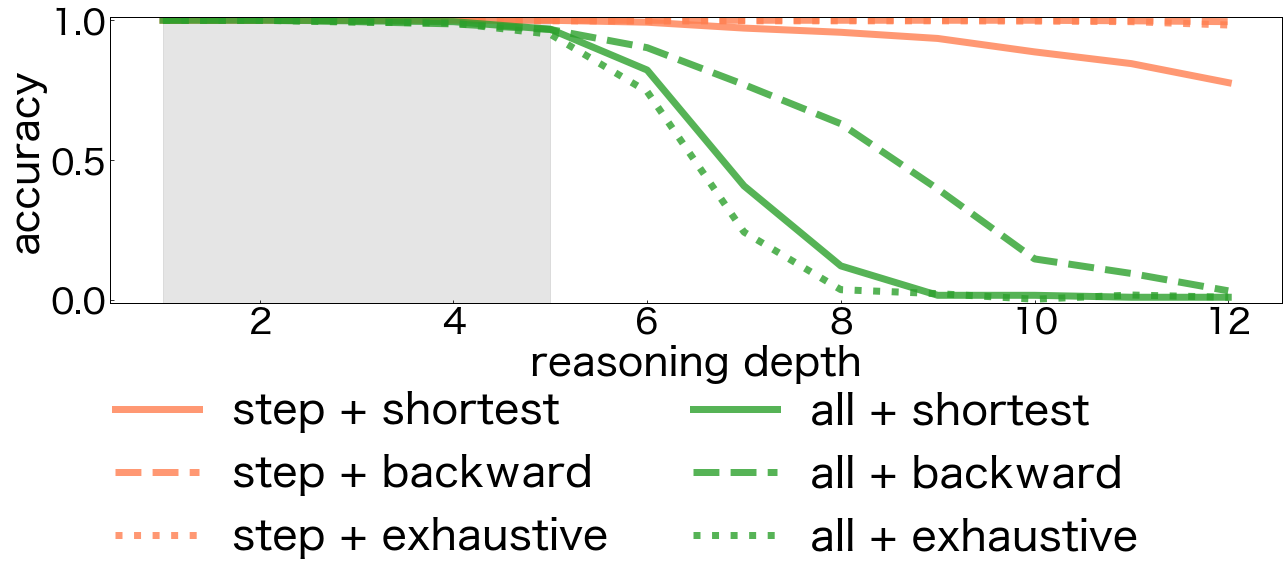}
\caption{Chaining strategy}
        \label{fig:result-chaining}
    \end{subfigure}
    %\caption{Overview of \textcolor{outputblue}{output} (left) and \textcolor{chainingpink}{chaining} (right) strategies.}
    \caption{Accuracy changes of the models against reasoning depth. The gray range represents the training data domain (1-5 depth). Figure (\ref{fig:result-output}) shows the performance degradation with the increase of reasoning steps when using the all-at-once strategy. Figure (\ref{fig:result-chaining}) shows that the combination of step-by-step output and backward/exhaustive chaining leads to successful generalization.}
    \label{fig:result}
\end{figure*}

\section{Results}
\label{experiment}
\label{subsec:models}
\paragraph{Models:}
We used the pre-trained T5-base, T5-large~\footnote{\url{https://huggingface.co/docs/transformers/model_doc/T5}}~\cite{article_t5}, and BART-base~\footnote{\url{https://huggingface.co/docs/transformers/model_doc/bart}}~\cite{article_bart}. Results of BART-base are in Appendix~\ref{sec:appendix_architecture}.

Note that their pre-defined tokenizers have all the numbers from \texttt{0} to \texttt{9}, and the numerical values in our dataset are divided into digits (e.g., ``\texttt{12}'' should be ``\texttt{@@1 @@2}'') in advance, following~\citet{Kim2021}.
%%%%%%%%%%%%%%%%%%%%%%%%%%%%%%%%%%%%%%%%%%%%%%%%

\noindent
\textbf{Training: }
The models were first pre-trained using a 10K \textit{simple} dataset for 30 epochs, then trained with the 5K training set (1K training instances for each reasoning depth.) for 2000 epochs. The experiment setting details are in Appendix~\ref{sec:appendix_experiment}.
In addition, we prepared 0.2K test instances for each reasoning depth. 
This pre-training is intended to teach the models primitive operations (i.e., assignment, reference, and addition).
The pre-training dataset contains two types of single-depth instances: \textit{assign-refer} type (e.g., \texttt{A=1,A?}) and \textit{operate-assign-refer} type (e.g., \texttt{A=1+3, A?}). 
 All the results in the paper are averages of the results on three different seeds. 
%%%%%%%%%%%%%%%%%%%%%%%%%%%%%%%%%%%%%%%%%%%%%%%%
\subsection{Output strategies}
\label{experiment_decoding_methods}
We compared the output strategies while fixing the chaining strategy to the shortest path.
Figure~\ref{fig:result-output} shows the accuracy per reasoning depth. 
Note that the accuracy score here denotes whether the answer (e.g., \texttt{C=6}) is correct.
We observed the following: (i) \textbf{generating intermediate reasoning steps enhance the performance}, and (ii) among the output strategies, \textbf{step-by-step works the best}, and \textbf{all-at-once works the worst}.
The format of the dataset in this study is simple. Therefore, this result indicates the low symbolic reasoning ability of neural models and the necessity of the choice of an appropriate reasoning strategy. 

We hypothesized that the source of all-at-once's inferiority was that the decoder overfitted to output a similar length of reasoning steps as those in the (shallower) training data. 
In fact, the models generated relatively shorter reasoning steps in the out-of-domain (e.g., depth of 12) setting when using the all-at-once strategy (Figure~\ref{fig:hist}); this supports our hypothesis.

The advantage of step-by-step over token-by-token suggests the advantage of breaking the problem into meaningful units (reasoning step) and modeling each step in a single call of the encoder-decoder.
%%%%%%%%%%%%%%%%%%%%%%%%%%%%%%%%%%%%%%
\begin{table}[t]
\centering
\footnotesize
\begin{tabular}{cp{1.6cm}p{1.3cm}p{1.3cm}}
\toprule
Depth & \multicolumn{1}{r}{Shortest} & \multicolumn{1}{r}{Backward} & \multicolumn{1}{r}{Exhaustive} \\
\cmidrule(r){1-1} \cmidrule(lr){2-2} \cmidrule(lr){3-3} \cmidrule(l){4-4}
6 & \multicolumn{1}{r}{99.3/99.3} & \multicolumn{1}{r}{100/\, 100} & \multicolumn{1}{r}{99.7/99.7} \\
8 & \multicolumn{1}{r}{95.5/95.7} & \multicolumn{1}{r}{100/\, 100} & \multicolumn{1}{r}{99.8/99.8} \\
12 & \multicolumn{1}{r}{76.7/77.7} & \multicolumn{1}{r}{99.5/99.5} & \multicolumn{1}{r}{98.2/98.3}\\
\bottomrule
\end{tabular}
\caption{Accuracy of the T5-base model with the step-by-step output strategy at each depth (chain/answer).}
\label{table:mirror}
\end{table}
\begin{table}[t]
\centering
\small
\begin{tabular}{lp{1.5 cm}p{1.5 cm}}
\multicolumn{3}{l}{Question: \texttt{A=1, B=2+A, B?}} \\
\toprule
Error types & Gold & Prediction \\ 
\cmidrule(lr){1-1} \cmidrule(lr){2-2} \cmidrule(l){3-3}
Copying error& \texttt{\textbf{B=2+A},} & \texttt{\textbf{B=6+A},} \\
& \texttt{B=2+1,} & \texttt{B=6+1,}\\
& \texttt{B=3} & \texttt{B=7}\\
\cmidrule(lr){1-1} \cmidrule(lr){2-2} \cmidrule(l){3-3}
Hasty assignment& \texttt{\textbf{B=2+A},} & \textcolor{gray}{(skip step)}\\
&\texttt{B=2+1,}& \texttt{B=2+2,}\\
& \texttt{B=3} & \texttt{B=4}\\
\bottomrule
\end{tabular}
\caption{Illustrative examples of the errors under the step-by-step, shortest-path chaining settings. \textcolor{gray}{(skip step)} denotes that the reasoning steps is accidentally skipped.}
\label{table:error}
\end{table}
%%%%%%%%%%%%%%%%%%%%%%%%%%%%%%%%%%%%%%
\subsection{Chaining strategies}
\label{experiment_reasoning_step}
Figure~\ref{fig:result-chaining} and Table~\ref{table:mirror} show the results on each depth with a fixed step-by-step output strategy. Note that the accuracy of the chain (left side of the scores) was measured based on not an exact match but mathematically. For example, even if the order of generated equations is different, it is correct. The results of a fixed token-by-token output strategy are in Appendix~\ref{sec:appendix_token_by_token}.

While the performance dropped in the shortest-path setting as the reasoning depth increased, with either the exhaustive or backward chaining, models successfully solved the task even when extrapolating to depths 6-12. 
The models correctly generated the intermediate steps (nearly perfect) as well as the final answer in the exhaustive and backward chaining settings (Table~\ref{table:mirror}).
Note that these strategies were ineffective with all-at-once outputting.

\citet{Gontier2020} compared chaining strategies and concluded that models that \textit{didn't} generate reasoning steps had better generalization performance than models that did when the reasoning chains were long. However, our results suggest that the choice of the appropriate output strategy improves the reasoning ability of the model.

We considered that the source of shortest-path inferiority was the rough granularity of the given reasoning steps. 
The models don't know the shortest path before outputting the reasoning steps. Therefore, both the exhaustive and shortest path chaining approaches must search for variables other than those on the shortest path. 
As shown in Figure~\ref{fig:chaining-strategies}, the exhaustive chaining approach is taught this process explicitly.
On the other hand, the shortest-path chaining approach must be learned that by training data that don't include this process. 
We thought this difference affected the accuracy and concluded that \textbf{the accuracy is higher when the granularity of given intermediate steps is finer}, even though they are long.

Therefore, we concluded that \textbf{the accuracy is higher when the granularity of intermediate steps is finer}, even though they are long.
%%%%%%%%%%%%%%%%%%%%%%%%%%%%%%%%%%%%%%%%
\subsection{Error analysis}
We also analyzed the errors of the depth-12 instances under the shortest-path strategy.~\footnote{In total, 32 instances were analyzed. That is the total number of incorrect answers on one seed.}
We observed two types of errors: (i) copying errors and (ii) hasty assignment.
Table~\ref{table:error} shows an illustrative example of each error type and the percentage of these errors.
The most frequent one (53\%) was a simple copying error, where the model failed to accurately copy an original equation into the reasoning chain.
This erroneous copying ability is consistent with \citet{DBLP:conf/acl/XuLYWHZ20} and supports the advantage of introducing a copy mechanism to the model~\cite{Ontanon2022-gy}.
Second, a hasty assignment is the model skipping the step to copy the equation from context and instead assigned it a random value. 
Note that these errors were almost addressed in the other strategies; this could stem from the difficulty of the implicit calculation of the shortest path.
%%%%%%%%%%%%%%%%%%%%%%%%%%%%%%%%%%%%%%%%
\subsection{Models' scalability}
\begin{figure}[t]
\centering
\includegraphics[width=7.5cm]{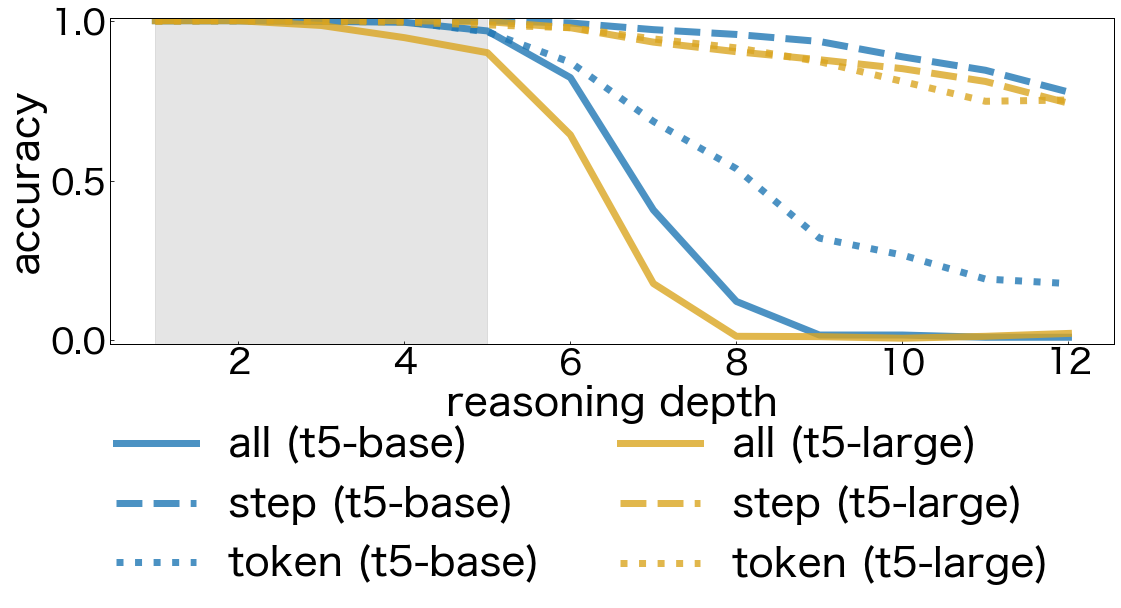}
\caption{Accuracy changes of the T5-base and T5-large against reasoning depth. The gray range presents the training data domain (1-5 depth). This figure shows that the accuracy of T5-large with token-by-token is higher.}
\label{fig:large}
\end{figure}

To investigate the scalability, we compared T5-large with T5-base. Figure~\ref{fig:large} shows the result.
T5-large had a similar trend but slightly lower accuracy on all-at-once and step-by-step compared to T5-base. The reason may be that T5-large needs more data for updating the weights of the entire model. On the other hand, the accuracy of T5-large is higher than T5-base on token-by-token. It's because the data size of token-by-token is as token lengths of output sequence times as the data size of all-at-once, as shown in Figure~\ref{fig:output-strategies}. This result indicates that the parameter size of the model needs to be larger to output token-by-token.
%%%%%%%%%%%%%%%%%%%%%%%%%%%%%%%%%%%%%%

%%%%%%%%%%%%%%%%%%%%%%%%%%%%%%%%%%%%%%
\section{Conclusions}
We investigated and factorized the reasoning strategy in symbolic numerical reasoning with neural seq2seq models.
We found that the combination of step-by-step output and finely granular reasoning leads to successfully performing symbolic reasoning. Our results support the potential of neural models for symbolic reasoning.
%%%%%%%%%%%%%%%%%%%%%%%%%%%%%%%%%%%%%%
\section*{Limitations}
We found that even simple symbolic reasoning requires the appropriate selection of reasoning strategy. 
It is unclear whether our findings generalize to more complex symbolic reasoning and/or problems written in natural language. 
If our findings do not generalize in these different settings, we must address the gap in future work. For example, we start with one of the simplest tasks and find out when models fail as we add complexity to tasks one by one. 

From the engineering perspective, the iterative strategies are limited to the input length of the model.
For example, in our experiments, when adopting the setting where reasoning depths are greater than 13, the input length of step-by-step and token-by-token became longer than the input length limit of T5 (i.e., 512 tokens).

In addition, gigantic language models (e.g., GPT-3) have recently been used. 
Including these models in our study is one of our future works.
%%%%%%%%%%%%%%%%%%%%%%%%%%%%%%%%%%%%%%
\section*{Acknowledgements}
We thank four anonymous reviewers who provided valuable feedback. We would like to also appreciate the member of Tohoku NLP Group for their cooperation in conducting this research.

This work was supported by JSPS KAKENHI Grant Numbers JP22H00524, 21K21343 and JST CREST Grant Number JPMJCR20D2, Japan.
%%%%%%%%%%%%%%%%%%%%%%%%%%%%%%%%%%%%%%%%%%%%%%%%
\bibliography{custom}
\bibliographystyle{acl_natbib}
%%%%%%%%%%%%%%%%%%%%%%%%%%%%%%%%%%%%%%%%%%%%%%%%
\clearpage
\appendix
\begin{figure}[t]
\centering
\includegraphics[width=7.5cm]{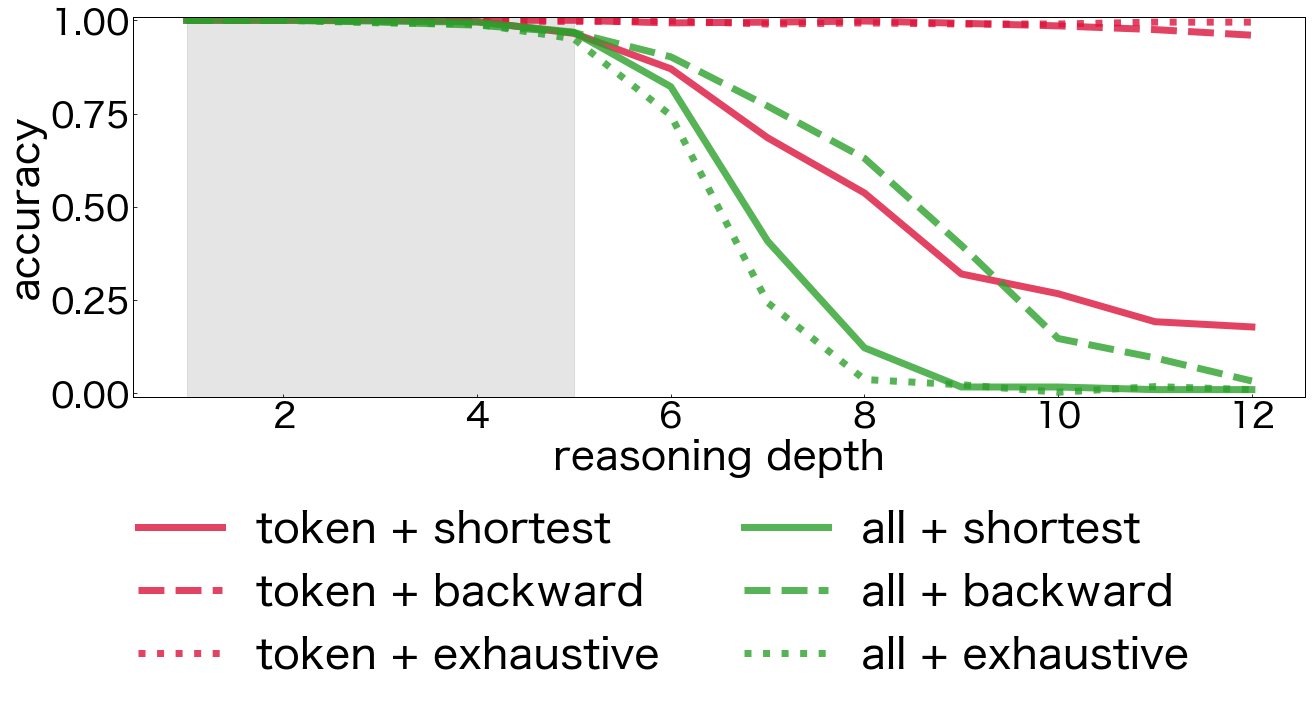}
\caption{Accuracy changes of token-by-token per reasoning depth. The gray range presents the training data domain (depths 1-5).}
\label{fig:token}
\end{figure}
%%%%%%%%%%%%%%%%%%%%%%%%%%%%%%%%%%%%%%
\begin{figure}[t]
\centering
\includegraphics[width=7.5cm]{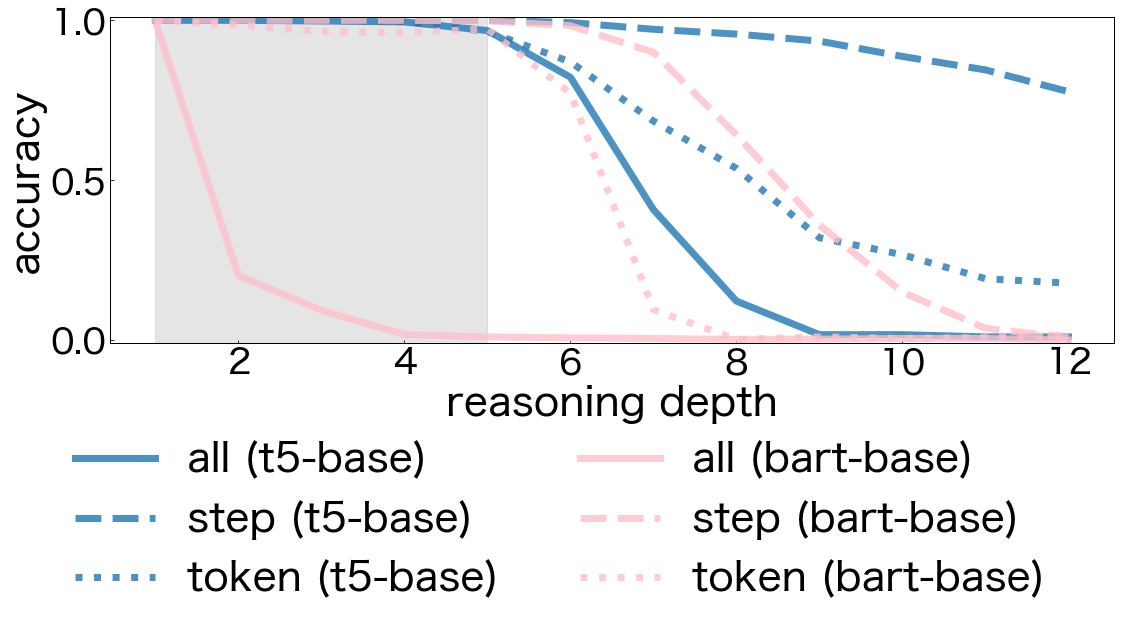}
\caption{Accuracy changes of the T5-base and BART-base models per reasoning depth. The gray range presents the training data domain (depths 1-5). T5 seems to outperform BART. }
\label{fig:bart}
\end{figure}
%%%%%%%%%%%%%%%%%%%%%%%%%%%%%%%%%%%%%%%%
\label{sec:appendix}
\section{Details on Experimental Settings}
\label{sec:appendix_experiment}
We first examined the learning rate from $10^{-3}$, $10^{-4}$, and $10^{-5}$; among them, we used the largest rate at which the loss converged. 
After training models, we used the model with the lowest validation loss among the per-epoch checkpoints during the training reported. 
We used four NVIDIA V100 GPUs for NVLink 16GiB HBM2.
%%%%%%%%%%%%%%%%%%%%%%%%%%%%%%%%%%%%%%
\section{Results of Token-by-token}
\label{sec:appendix_token_by_token}
Figure~\ref{fig:token} shows the results on each depth with a fixed token-by-token output strategy. Like step-by-step, the performance drops in the shortest-path setting as the reasoning depth increases. In addition, the exhaustive or backward successfully solves the task even when extrapolating to depths 6-12. 
%%%%%%%%%%%%%%%%%%%%%%%%%%%%%%%%%%%%%%
\section{Different Architectures}
\label{sec:appendix_architecture}
We also tested BART-base~\cite{article_bart} as a baseline to investigate the effectiveness of the NLP-task-oriented objectives used in the T5-style pre-training. Figure~\ref{fig:bart} shows this result. In this particular setting, T5 was superior to BART. This suggests that the NLP-task-oriented objectives benefit symbolic reasoning.
%%%%%%%%%%%%%%%%%%%%%%%%%%%%%%%%%%%%%%
\begin{table*}[t]
\centering
\small
\begin{tabular}{lp{5.5 cm}p{5.5 cm}}
\multicolumn{3}{l}{Question: \texttt{A=1, C=5+B, B=2+A, D=3+A, C?}} \\
\toprule
Chain error types & Gold & Prediction \\ 
\cmidrule(lr){1-1} \cmidrule(lr){2-2} \cmidrule(l){3-3}
Ignoring the incorrect step & \texttt{A=1, \textbf{B=2+A}, B=2+1, B=3, C=5+B, C=5+3, C=8}& \texttt{A=1, \textbf{B=2+D, B=2+A}, B=2+1, B=3, C=5+B, C=5+3, C=8} \\ 
\cmidrule(lr){1-1} \cmidrule(lr){2-2} \cmidrule(l){3-3}
Correct assignment & \texttt{A=1, \textbf{B=2+A, B=2+1}, B=3, C=5+B, C=5+3, C=8} & \texttt{A=1, \textbf{B=2+D, B=2+1}, B=3, C=5+B, C=5+3, C=8} \\
\cmidrule(lr){1-1} \cmidrule(lr){2-2} \cmidrule(l){3-3}
Non affecting error & \texttt{A=1, B=2+A, B=2+1, B=3, C=5+B, C=5+3, C=8} & \texttt{A=1, B=2+A, B=2+1, B=3, \textbf{D=3+A, D=3+2, D=5}, C=5+B, C=5+3, C=8}  \\ 
\bottomrule
\end{tabular}
\caption{These instances are examples of chain errors. Note that the final answers are correct.}
\label{table:chain_error}
\end{table*}
%%%%%%%%%%%%%%%%%%%%%%%%%%%%%%%%%%%%%%%%%%%%%%%%%%%%%%%%%%%%%%%%%%%%%
\section{Other errors}
\label{sec:appendix_error}
We analyzed the cases where the answer is correct and the chain is wrong. Table~\ref{table:chain_error} shows examples of chain errors. Ignoring the incorrect step is an example of the model outputting the correct reasoning step after outputting an incorrect one. Correct assignment is an example in which the assignment accidentally makes the model output the correct step. Finally, Non-affecting error is an example in which a variable not on the shortest path is wrongly assigned a value.

\end{document}